\documentclass[10pt,twocolumn,letterpaper]{article}

\usepackage{cvpr}
\usepackage{times}
\usepackage{epsfig}
\usepackage{graphicx}
\usepackage{amsmath}
\usepackage{amssymb}
\usepackage{subcaption}
\usepackage{float}
\usepackage{array}
\usepackage{multirow}
\usepackage{hhline}

\usepackage[pagebackref=true,breaklinks=true,letterpaper=true,colorlinks,bookmarks=false]{hyperref}

\cvprfinalcopy 

\def\cvprPaperID{???} 

\ifcvprfinal\fi
\begin{document}

\title{CoMaL Tracking: Tracking Points at the Object Boundaries}

\author{Santhosh K. Ramakrishnan$^{1}$ \quad
Swarna Kamlam Ravindran$^{2}$ \quad
Anurag Mittal$^1$ \vspace{0.4em}\\
$^1$IIT Madras \quad \quad \quad $^2$Duke University \quad \quad \quad\vspace{0.2em} \\
{\tt\small \{ee12b101@ee, amittal@cse\}.iitm.ac.in, swarnakr@cs.duke.edu}
}

\maketitle

\begin{abstract}

Traditional point tracking algorithms such as the KLT use local 2D information aggregation for feature detection and tracking, due to which their performance degrades at the object boundaries that separate multiple objects.  Recently, CoMaL Features have been proposed that handle such a case.  However, they proposed a simple tracking framework where the points are re-detected in each frame and matched.  This is inefficient and may also lose many points that are not re-detected in the next frame.  We propose a novel tracking algorithm to accurately and efficiently track CoMaL points.  For this, the level line segment associated with the CoMaL points is matched to MSER segments in the next frame using shape-based matching and the matches are further filtered using texture-based matching.  Experiments show improvements over a simple re-detect-and-match framework as well as KLT in terms of speed/accuracy on different real-world applications, especially at the object boundaries. 

\end{abstract}

\section{Introduction}
\label{sec:Introduction}

Feature Point Detection, Matching and Tracking is an important problem that has been studied extensively in the Computer Vision literature and has numerous applications such as Mosaicing, Object Tracking~\cite{song2014vehicle, tanathong2014translation, cao2011klt}, Action Recognition~\cite{uemura2008feature, matikainen2009trajectons, messing2009activity, wang2011action} and Structure-from-Motion~\cite{akbarzadeh2006towards, bok2007accurate, pollefeys2008detailed} among others.
The Kanade-Lukas-Tomasi (KLT)~\cite{lucas1981iterative, tomasi1991detection, shi1994good} tracker is still the most widely used tracker in the literature even after 30 years due to its robustness and speed.  In KLT, Harris corners~\cite{harris_1988_tracking} are detected in the first frame and are subsequently tracked using iterative search of the matching image patch around the detected point using a gradient descent approach.  Several extensions to the original KLT have been proposed.  For instance, ~\cite{baker2001equivalence} proposes several variations of the original KLT algorithm, while~\cite{baker2004lucas} improves its efficiency.  GPU-based extensions~\cite{sinha2006gpu, zach2008fast} of KLT have also been proposed to obtain significant speed-ups over the traditional implementations. 

\begin{figure}[t]
\begin{center}
   \includegraphics[width=1.0\linewidth, trim=0 220 340 20, clip]{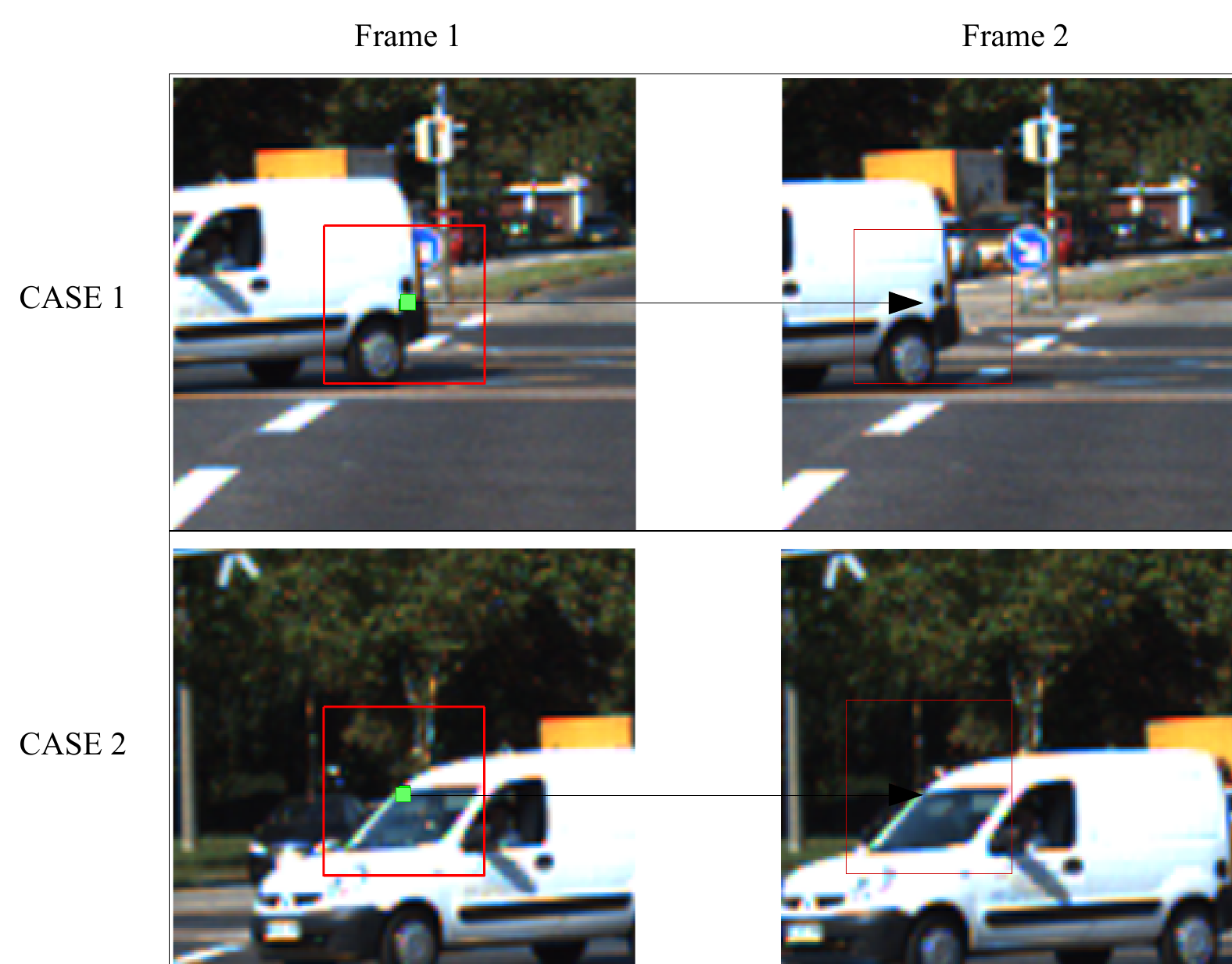}
\end{center}
\caption{Case of KLT point tracking failure at the object boundary due to a large change in the background portion of the support region.}
\label{fig:klt-failures}
\vspace{-2em}
\end{figure}

While KLT has been the state-of-the art for feature point tracking, other methods have also been proposed~\cite{bolme2010visual, danelljan2016beyond} and can be used for Feature Point Detection and Tracking.  
However,  almost all these methods including the KLT work well only in the interior of objects and do not perform very well at the object boundaries.  This is due to the consideration of a full 2D support region around a point for matching which can be problematic at the object boundaries where the background portion of the support region can change.  This is illustrated in Figure~\ref{fig:klt-failures}. 

There have been other algorithms to address the issue of varying backgrounds in the boundary regions of objects.  Mikolajczyk \etal~\cite{mikolajczyk2003shape} use edge-based features and identify the dominant edge to separate the two regions at the object boundary for the problem of Object Recognition.  For the task of Object Tracking, SegTrack~\cite{almomani2013segtrack}, Chen \etal~\cite{chen2013constructing} and Oron \etal~\cite{oron2014extended} iteratively build foreground and background appearance models for robust object tracking. However, these methods degrade in performance when the object boundaries dominate the object appearance (as in the case of thin objects) and require a good initilization to iteratively segment the foreground and the background. 

The CoMaL Point Detector~\cite{Ravindran_2016_CVPR} has been proposed recently, which addresses many of such issues at the Object oundaries without the need for a good initialization and an iterative approach.  It is based on the idea of level lines that often separate the foreground from the background and are fairly robust to illumination changes happening on one side of the divide.  Furthermore, CoMaL Point Matching allows for matching only one of the two sides of the level line, thus making it invariant to a change on one side due to a background change.  
It has also been observed in the literature that Maximally Stable Extremal Regions (MSERs)~\cite{matas2004robust}, the seminal work that  originally used such level lines for Point Detection, are quite robust compared to other corner points since they are invariant to an affine transformation of image intensities~\cite{matas2004robust, mikolajczyk2005comparison}, and were also found to be extremely stable~\cite{matas2004robust} and highly repeatable in many comparative studies~\cite{mikolajczyk2004comparison, mikolajczyk2005comparison}. 

Although the CoMaL features are very good for the case of stable Feature Point Detection and Re-detection/Matching at the Object boundaries, the problem of tracking in continuous videos remained unaddressed, although this can be done naively by re-detecting all points in the next frame and matching.  However, such an approach will fail if the corresponding feature point does not get detected in the next frame.  Furthermore, feature point detection for each frame is an expensive step and reduces the efficiency of tracking. We propose an alternate algorithm for tracking CoMaL points with several contributions. First, instead of re-detecting the points again for each frame which is an expensive step, we search for the corners present in the previous frame in some given neighborhood.  This makes it not only efficient but also alleviates the problem of missed corner point detections.  Second, in order to do such a search, we first do a shape-based matching of the level line segment associated with a given corner point in the neighborhood.  Such a matching is done on the MSER boundaries found in the next image and not on the edge map, which makes the method quite robust.  Third, as in the original CoMaL work~\cite{Ravindran_2016_CVPR}, we further filter such matches by doing an SSD matching on one side of the CoMaL level line.  All these steps are robust to  changes on one side of the level line and yield a method for tracking CoMaL points that works reliably and efficiently at object boundaries. 

We first give a review of the CoMaL Point Detector~\cite{Ravindran_2016_CVPR}, which are used as a base for our tracker.

\section{The CoMaL Corners}
	
The CoMaL Feature Point Detector~\cite{Ravindran_2016_CVPR} identifies corners on iso-intensity curves or level lines.  Such level lines have been found to be fairly stable under many image transformations and have been used as a base for several Feature Point detectors such as MSER~\cite{matas2004robust} and CoMaL.  They can also be reliably detected at the object boundaries, which they often trace.  The CoMaL Corners are identified as the points of high curvature on stable portions of long level lines, i.e., a corner point must satisfy two conditions: (a)  It must lie on a stable level line segment and (b)  have a high ``cornerness" value at a given scale. The stability of a level line segment is inversely proportional to the area between it's two neighbouring level line segments and signifies the motion of the level line upon a certain change in the intensity.  The cornerness measure is defined based on the eigen values of the point distribution centered around the corner at a particular scale on the level line and large eigen values in both directions signify a ``turn" of the level line at that point and hence a corner point.

\begin{figure}[t]
\begin{center}
   \includegraphics[width=1.0\linewidth, trim=20 430 390 10, clip]{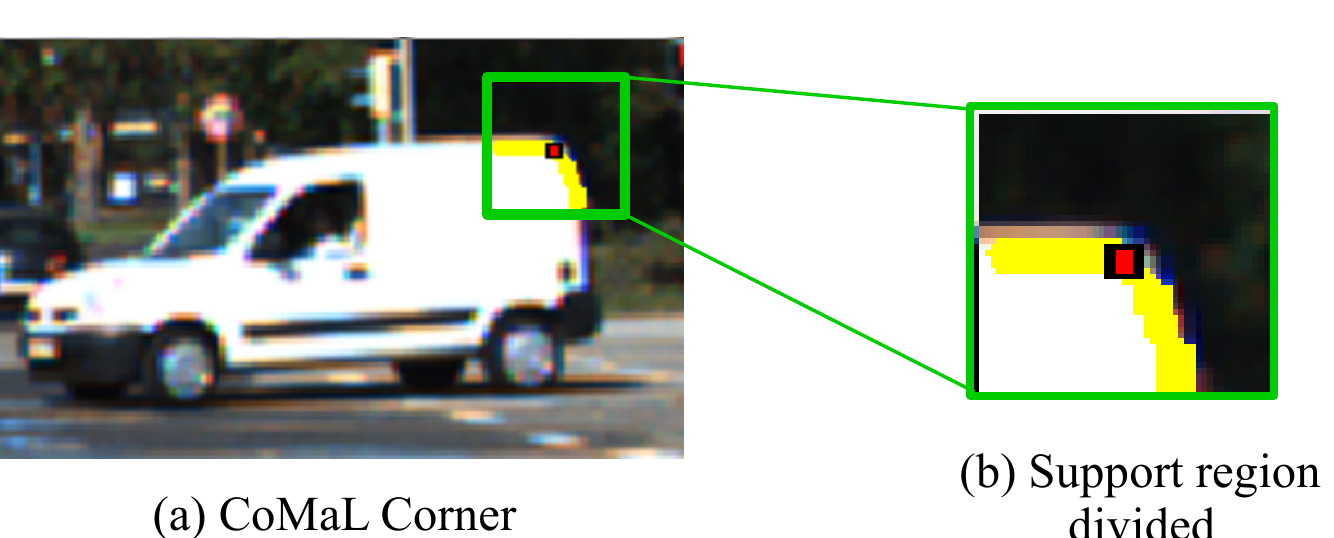}
\end{center}
   \caption{(a) Example of CoMaL Corner Point(red) and associated level line(yellow). (b) The support region of the corner (green box) divided into the regions belonging to the two objects (foreground and background segments) by it's level line segment. }
\label{fig:comal}
\end{figure}

CoMaL points were shown to be more reliable and stable on the object boundaries compared to other feature point detectors such as FAST~\cite{rosten_2005_tracking}, Harris~\cite{harris_1988_tracking}, Hessian~\cite{mikolajczyk2004scale} and MSER~\cite{matas2004robust}, and comparable to them in the interior of objects in the original CoMaL paper~\cite{Ravindran_2016_CVPR}. Also, the paper developed a reliable approach for matching corners at object boundaries by dividing the support region of the corner by the CoMaL level line into two regions, as shown in Figure~\ref{fig:comal}. By independently matching the two regions, it allows us to compute a part SSD score by matching only one part of the support regions of the two corner points.  Thus, if there is a change due to a background change in one of the parts, it can be neglected.  As a result, it allows robust matching of feature points across images even where the background may not be fixed.  Due to these characteristics of the CoMaL Feature Point Detector, such points are quite suitable for being tracked reliably at the object boundaries.  How we do so is described in the next section.

\section{The Tracking Algorithm}
\label{sec:tracking_algorithm}

\begin{figure}[t]
\centering
\includegraphics[width=1.0\linewidth, trim=70 30 300 30, clip]{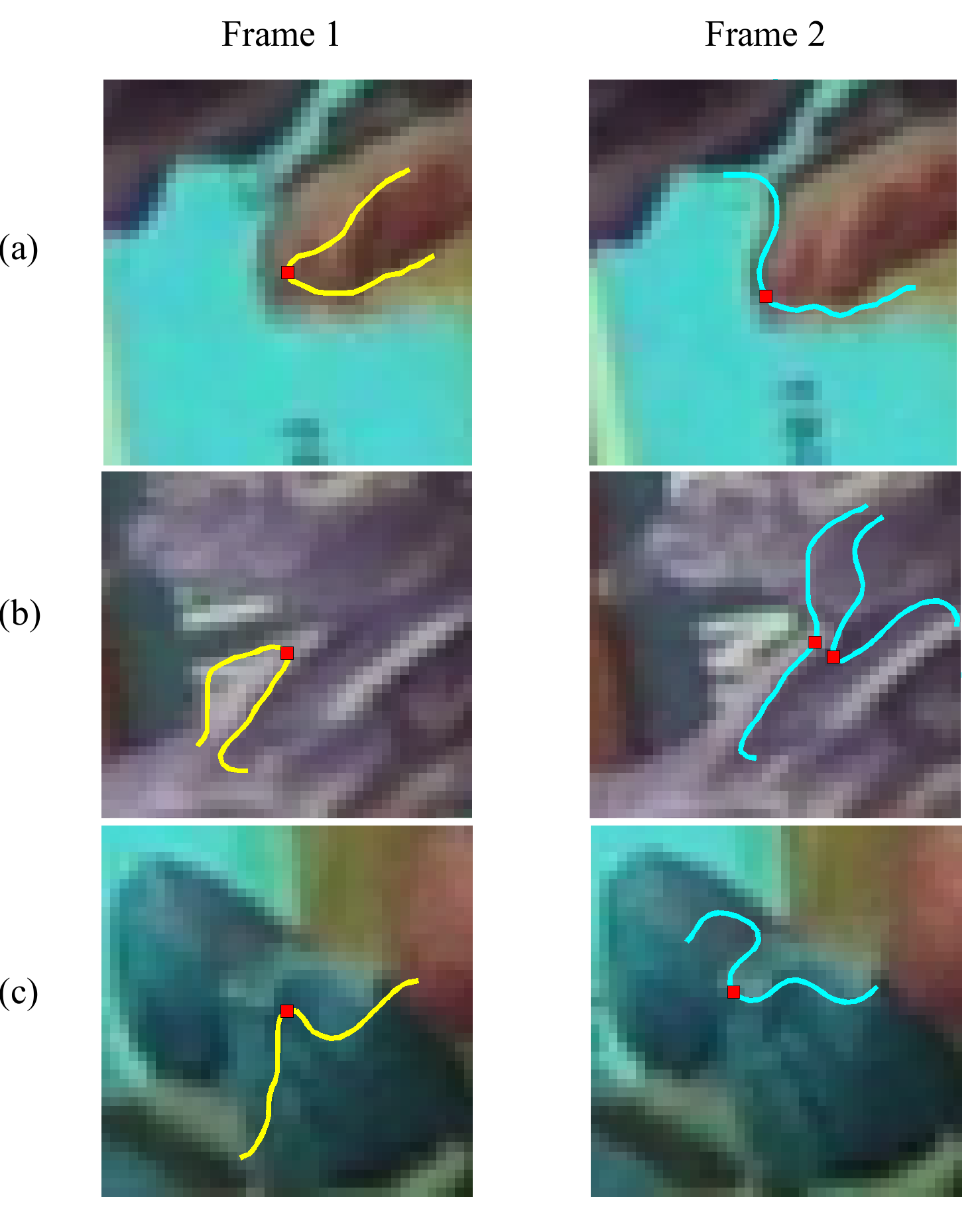}
\caption{Three failure cases of CoMaL re-detect-and-match framework. Corner (red) to be tracked shown in first frame with associated level line (yellow). Detected corners (red) shown in second frame with associated level lines (blue). No matches are found in frame 2 for the corner in frame 1.}
\vspace{-1em}
\label{fig:tbd-failures}
\end{figure}

The original CoMaL Point Detector paper~\cite{Ravindran_2016_CVPR} presents a method for matching points across frames.  This method can be used for tracking as well by simply re-detecting points in the next frame and matching to the points in the previous frame.  However, apart from being slow, the re-detect-and-match method can fail if the corner in the next frame is not detected at the same position. This can happen if the corresponding point in the next frame falls below the cornerness threshold due to minor object deformations, illumination changes or if the corresponding level line segment was not stable in the next frame.  Examples are shown in Figure~\ref{fig:tbd-failures}. As we can see, the CoMaL Detector does not detect the corresponding corner in the second frame in the three cases shown. If we look at the cases more closely, corner detection could have failed if the corresponding level line segment was not maximally stable or if the corresponding point was not identified as a corner on the stable level line segment.  Due to such missed points, the given point will not be tracked correctly in the next frame.  Furthermore, this method is slow due to the high computational cost of point re-detection.  In this section, we describe a more efficient algorithm for tracking points across frames. 

\begin{figure}[t]
\centering
\includegraphics[width=1.0\linewidth, trim=100 210 400 120, clip]{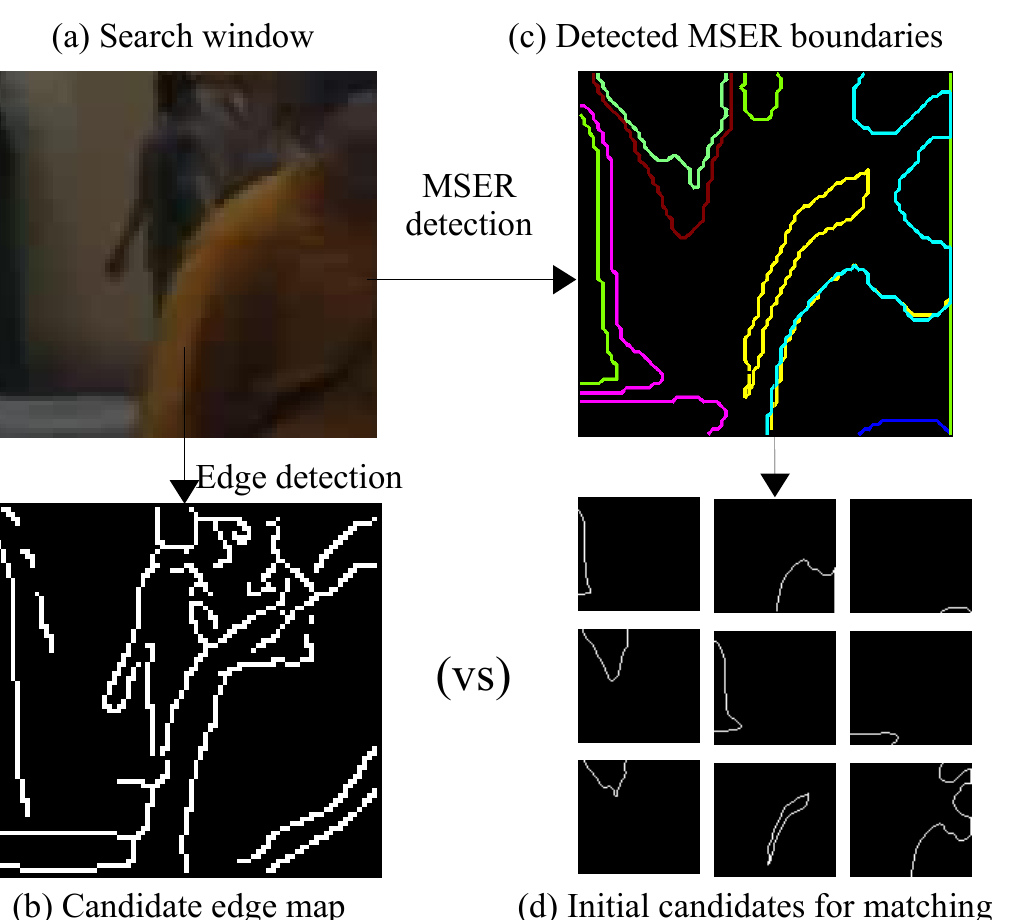}
\caption{Advantage of matching stable level lines over matching edges directly: (a) A sample search window. (b) The corresponding edge map. (c) The detected local MSER boundaries (each in a different color), (d) Each MSER boundary is matched individually to a CoMaL level line.}
\vspace{-1em}
\label{fig:mser_edge}
\end{figure}

We first try to track the level lines associated with the CoMaL corners.  The full corner patch cannot be tracked as a portion of the patch may have changed due to a background change.  Contour matching techniques can be used to track the level lines segments by matching them with edges in the search region.  However, the problem with this approach is that they can be over-segmented and broken due to loss of gradients along some portion of the level line as shown in Figure~\ref{fig:mser_edge}.  Furthermore, the level line segment can match to edges belonging to multiple level lines in the current image, which can lead to an erroneous match.  To address this issue, we only match the given level line segment with the individual stable level lines in the current frame, which can be  obtained easily using MSER boundary segments (Figure~\ref{fig:mser_edge}).  Since we know that CoMaL corners lie on stable level lines, matching the corner's level line segment with locally stable level line segments in the next frame is a more compatible way to match than matching them with edges in the next frame since we would be searching for only stable level lines in the current frame that are similar in shape to the CoMaL level line.  In order to account for possible loss of strength of the level line stability, the stable level lines are extracted with a lower threshold than is done in the CoMaL point detector (The detector needs to have a higher threshold so as to not detect weak corners, but we can afford it since we are only searching for the corner that was already detected in a previous frame.).

\begin{figure*}

\centering
\includegraphics[width=1\linewidth, trim=0 440 40 15, clip]{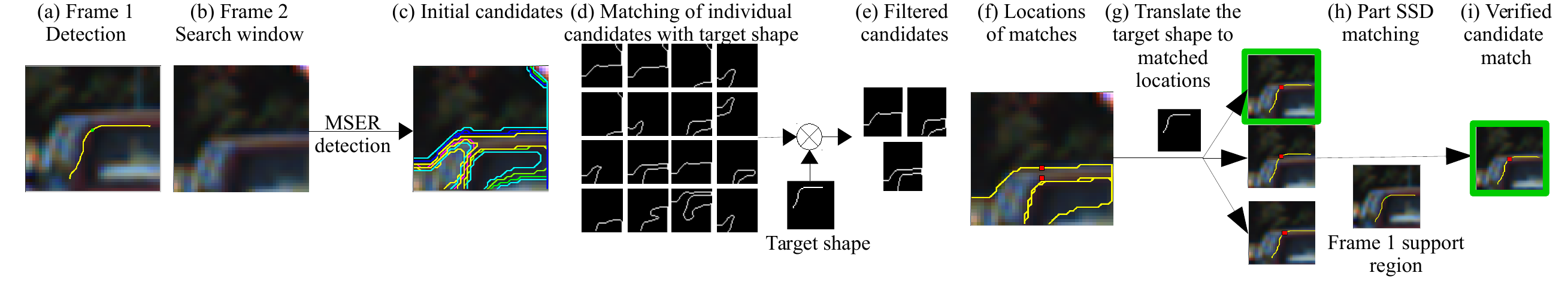}

\caption{Summary of Tracking Pipeline: We track the corner by tracking the associated level line. (a)-(e) Candidate shortlisting using shape matching (Section \ref{sec:shortlist} ). (f)-(i) Candidate verification using Part SSD Matching (Section \ref{sec:verification}) .}
\label{fig:tracking-pipeline}
\end{figure*}

Once the matching stable level lines are shortlisted by shape matching, the matching points are further verified by part SSD patch matching (matching only one side of the level line) as in the CoMaL matcher to screen out any false matches in the first stage.  
Our tracking algorithm can thus be divided into two phases (Figure~\ref{fig:tracking-pipeline}): \vspace{-0.7em}
\begin{enumerate}
	\item Shortlisting candidate matches using shape-based matching of stable level-line matches \vspace{-0.7em}
	\item Verification of filtered candidates using part SSD Matching
\end{enumerate}
\vspace{-1em}

\subsection{Shortlisting Candidates using Shape Matching}
\label{sec:shortlist}

\begin{figure}[t]
	\centering
	\includegraphics[width=1.0\linewidth, trim=10 250 400 10, clip]{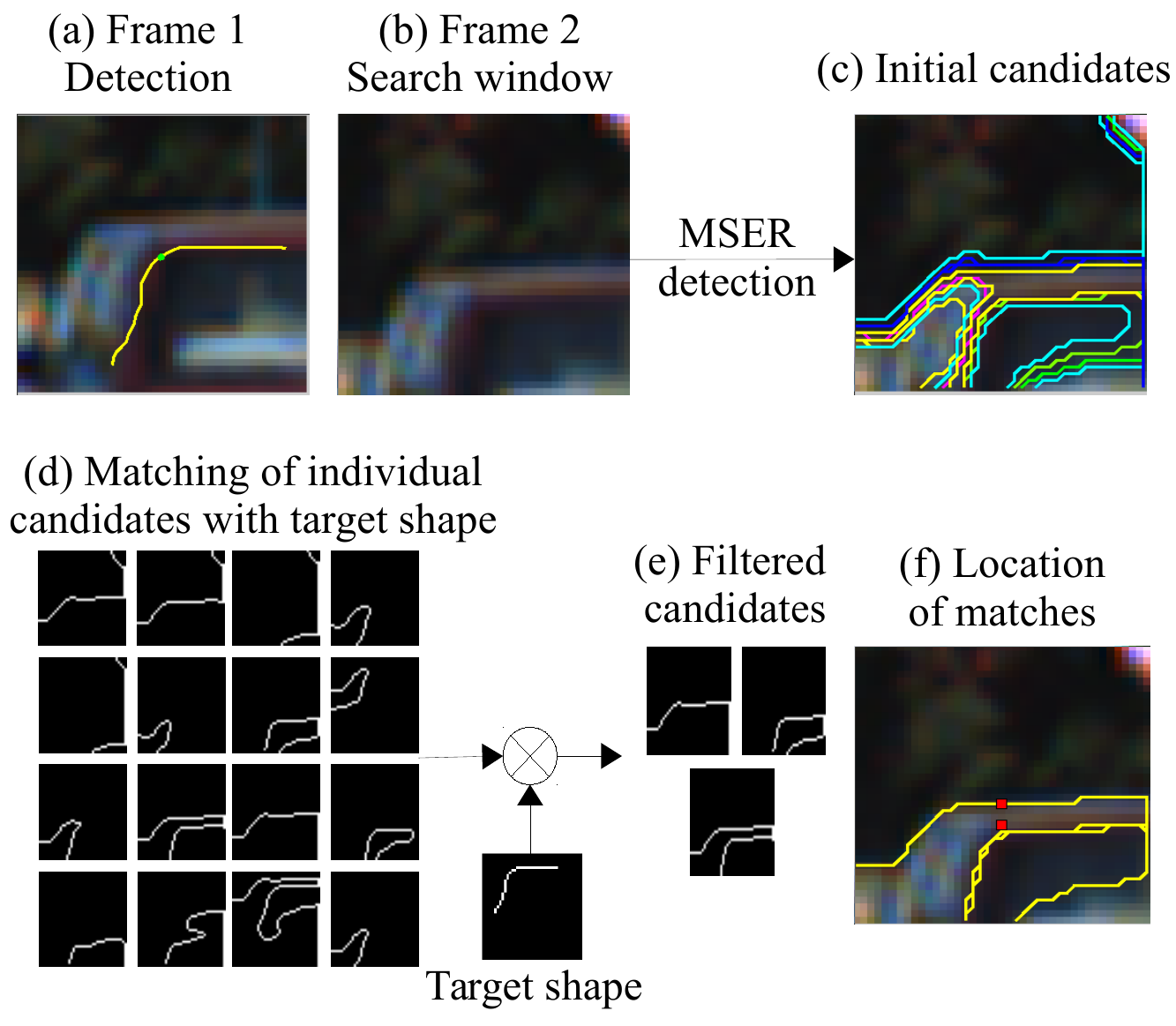}
	\caption{Candidates are shortlisted using shape matching. (a) The corner to be tracked in the first frame.  (b) The search region in the second frame. (c) Initial candidates obtained using MSER detection. (d) Candidates are individually matched using Hierarchical Chamfer Matching to the target shape to obtain filtered candidates as shown in (e). (f) The location of the matches in the search window are indicated in red.}
	\label{fig:pipeline-1}
\end{figure}

In order to shortlist candidate matches in the search region, we first perform an MSER~\cite{matas2004robust} detection in a local image patch and find stable local level line segments.  These, individually, form an initial set of target matching contour segments for the given CoMaL level line.  MSERs are detected in the search window and their boundary segments are obtained (Figure~\ref{fig:pipeline-1}~(c)).
(In practice, this step is speeded up by pre-computing MSERs in local image windows.  Then, for each point, we simply select the window closest to the search window and truncate it to the size needed.).  
As explained before, selecting MSER boundary segments as candidates for tracking is more compatible with the CoMaL corner detector compared to directly using edges because the CoMaL corners lie on level line segments. 
Next, we filter out the poorly matching candidates by computing a shape-based matching score between the level line segment of the corner and each candidate, and reject the candidates which have low matching scores. 
Note that this is done individually for each MSER segment separately.
Note also that the shape of the level line does not typically change even in the presence of a background change on one side of the level line, even as the level line might stride the object boundary.  Thus, this step can be done accurately even when the CoMaL point is at an object boundary.

We perform shape-based matching as shown in Figure~\ref{fig:pipeline-1}~(d). We have used Hierarchical Chamfer matching(HCMA)~\cite{borgefors1988hierarchical} to obtain a matching score between the candidates and the level line segment of the corner. Other matching methods could be potentially used in place of HCMA, depending on the requirements of the tracker, however, we use HCMA in our implementation because it is extremely fast and sufficient to obtain reliable matches across adjacent frames. In HCMA, matching begins at a low resolution and only the regions which were not rejected at lower resolutions are explored at higher resolutions.  The matching score is computed at the highest resolution using a Chamfer Matching criteria (average distance to the nearest edge point in the target image).

Most of the incorrect matches are filtered in this step, but a few matches are often left as shape is not fully discriminative.  Also, taking the best match by using only the shape criteria is sometimes not correct as the shape of the level line changes sometimes.  In order to select the best match, we next perform a texture-based verification step, the score of which is taken as the final score for selecting the best match.

\subsection{Match Verification using Part SSD Matching}
\label{sec:verification}

\begin{figure}[t]
	\centering
	\includegraphics[width=1.0\linewidth, trim=20 400 330 0, clip]{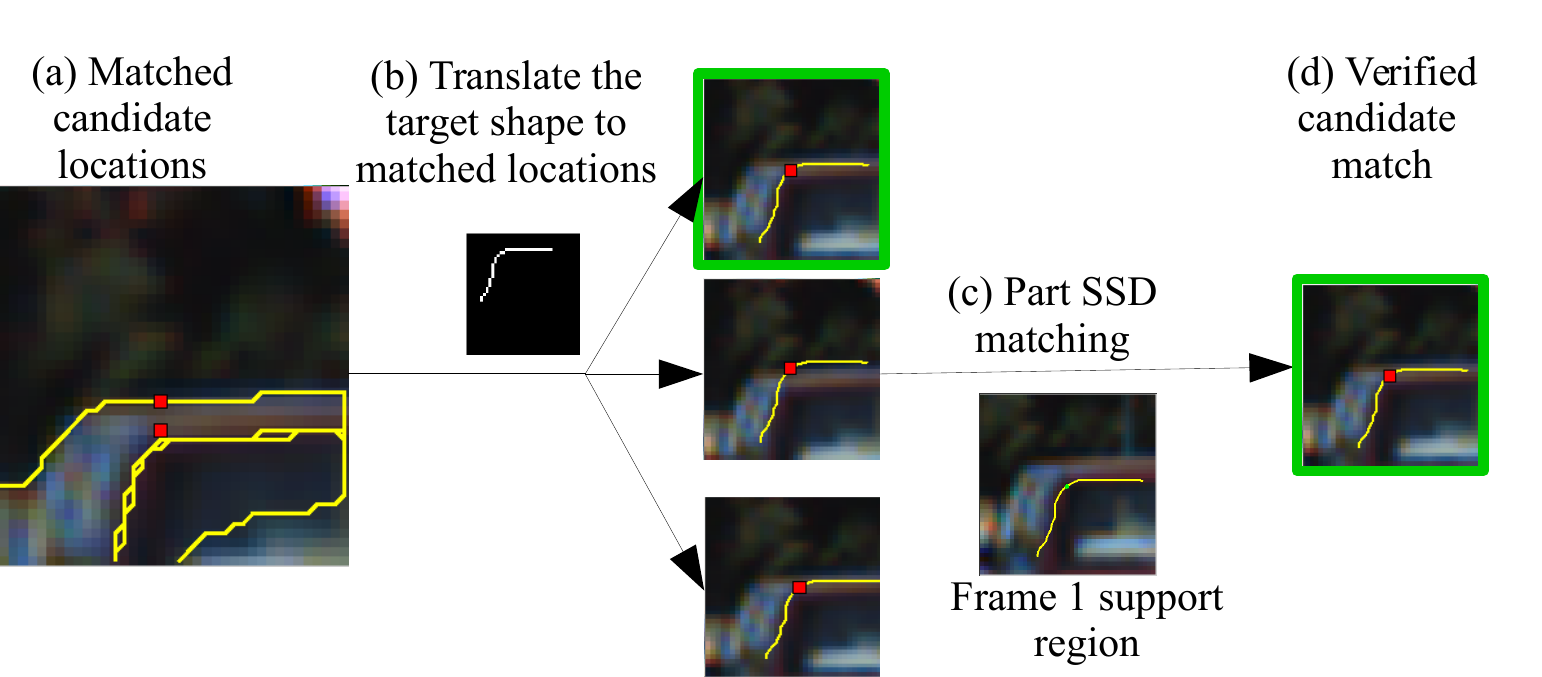}
	\caption{Candidate verification using texture based matching. (a) Location of matches of shortlisted candidates from~\ref{sec:shortlist}. (b) Target shape is translated to the matched locations and superimposed to obtain the final set of candidates. (c) The support region from frame 1 and the final candidates are matched using part SSD matching~\cite{Ravindran_2016_CVPR}. (d) The verified candidate match.}
	\label{fig:pipeline-3}
\end{figure}

Given the restricted set of candidates in the search window, we want to find the candidate that best matches with the CoMaL corner in the current frame.  We perform texture-based verification to select the best matching candidate among the filtered candidates.  In our algorithm, we use part SSD matching~\cite{Ravindran_2016_CVPR} to obtain the matching scores between the candidate MSER boundary segments and the level line segment of the original CoMaL corner. Part SSD matching independently matches the two parts of the support region divided by the level line, leading to four possible matching combinations for a given pair of candidate and corner level line segment. The best matching combination is selected and the corresponding score is  reported as the matching score between the pair. This is vital for tracking points on the object boundaries because the background keeps changing, so only the object portion of the support region can be reliably matched. As a result, this technique is better than a straight-forward full patch based SSD at the object boundaries.  Other sophisticated techniques such as HOG~\cite{dalal2005histograms}, normalized cross-correlation, SIFT~\cite{mikolajczyk2004scale}, \etc could potentially be used for more generalized matching scenarios, although they would have to be modified for part matching, which may not be easy.  Also, gradient-based matching may not be suitable for partial patch matching as only one side of the level line is used which may be homogenous.  Furthermore, these methods introduce invariances to certain transformations which may not be present in tracking applications, where there is limited variation across nearby frames.  This can unnecessarily introduce some false matches.  Thus, exact patch matching using SSD performs better in this scenario and is the basis for the KLT tracker as well.  The score obtained from such part SSD matching is used to select the best match as shown in Figure~\ref{fig:pipeline-3}.  
This two-stage selection process enables us to use both the shape and the texture information of the corner and its support region for matching and is thus fairly robust.

Our tracking algorithm is also more efficient than the re-detect-and-match framework.  This is because we do not perform the expensive step of corner detection in every frame.  Also, the part SSD matching algorithm used is more expensive because 4 independent matches have to be computed for each pair of corners as opposed to only one match for us (since the side that matches can be kept track of).  Also, we do not do an iterative optimization step for corner point optimization as in CoMaL point detection (this is not required as we are only tracking and not finding the corner afresh), and only do MSER detection once in local image overlapping patch segments.  All this is far less expensive than  full-blown CoMaL corner detection.  

\section{Experimental setup and Results}
\subsection{Baselines}

We consider two baseline algorithms for comparisons.

\subsubsection{The KLT Tracker}

The first baseline we use is the KLT tracker~\cite{lucas1981iterative}, which has been consistently used in applications such as Action Recognition~\cite{uemura2008feature, matikainen2009trajectons, messing2009activity, wang2011action}, Vehicle Tracking~\cite{song2014vehicle, tanathong2014translation, cao2011klt}, 3D Reconstruction~\cite{akbarzadeh2006towards, bok2007accurate, pollefeys2008detailed} in recent literature and is still the state-of-the-art even though it was proposed in 1981, suggesting it's effectiveness on a variety of applications.  As  explained in Section~\ref{sec:Introduction}, KLT may fail to track points on the object boundaries effectively as it uses the whole patch, which may not remain stable at the object boundaries.  We used the inverse compositional algorithm implementation of KLT in our experiments as this was shown to give the best results in the literature~\cite{baker2004lucas}.

\subsubsection{CoMaL Point Re-detect-and-Match Approach}

The second baseline we use is the original CoMaL paper's re-detect-and-match approach~\cite{Ravindran_2016_CVPR}, which was shown to perform better than other combinations of detectors and descriptors for detection and matching of feature points at the object boundaries.  While such re-detection and matching of features is essential after every $N$ frames due to lost points and will be essential in a complete system along with the tracking approach presented in this paper, 
this comparison serves to demonstrate the advantage of our algorithm over this simple re-detect-and-match strategy using the same detector and matcher.  We do not compare our results with other combinations of feature point detectors and descriptors since the CoMaL points were already shown to be much superior to the others for this task.  Also, such an approach does not do as well in general as a tracking approach used by trackers such as the KLT, which do not rely on a re-detection step which can miss some points, leading to a higher error while tracking.  Hence, comparisons with these other re-detect-and-match feature detectors and descriptors is not provided in this paper and the reader is referred to the original CoMaL paper~\cite{Ravindran_2016_CVPR} for such comparisons.

\subsection{The Evaluation Framework}
\label{sec:evaluation}

Since our datasets contain only object bounding box information and not exact point matching data, we generate the ground truth for point matching similar to~\cite{Ravindran_2016_CVPR}.  We assume that the relative location of a point w.r.t. the annotated bounding box remains the same across frames.  In order to account for non-rigidity of the objects and errors in the bounding box annotations, we allow small amounts of error between the ground truth location and the predicted match. The allowance given was 15 pixels in all the datasets. A common scale value of $8.4$ was selected for both the Harris and CoMaL detectors to allow for a fair comparison.  A support region of dimensions $41\times 41$ is used.  Since the precision-recall depends on the number of points generated by a detector using a threshold, we equalize the number of points generated by the point detectors on the different datasets.  The \textit{corner} function from MATLAB was used to obtain the Harris corner points
and the quality and sensitivity parameters were varied in order to obtain a varying number of Harris corner points.  Similarly, cornerness and stability threshold were varied for the CoMaL points as in the original paper.  

Following the evaluation protocol of~\cite{Ravindran_2016_CVPR}, the matching accuracy or precision is defined as the ratio of the number of correct matches to the total number of obtained matches.  Since the number of correct matches varies with the precision, as in \cite{Ravindran_2016_CVPR}, we report the number of correct matches obtained at a given precision, averaged over all the frames, to compare our algorithm with the baselines.  Our scores are reported as $\#matches/precision$ and a higher number of matches that are successfully tracked at the same precision indicate a better tracker.  If the total number of original points detected by the different detectors is the same (which is not always possible to achieve in practice), this also indicates a higher recall at the same precision.  We have chosen a typical operating precision value of around 0.8 for comparisons, although we had to decrease or increase this a bit to 0.7 or 0.9 for some sequences if the number of points was too little or too many at 0.8 precision.

\subsection{Results}

In our experiments, we show results on three different domains. In the first domain, we test our algorithm on a dataset for Object Tracking in a controlled setting that allows us to evaluate in detail the tracking performance on boundary $\&$ non-boundary regions for the different approaches. Next, we present results on Vehicle Tracking, which is a more realistic and critical application, but does not have much 
object rotation as in the first dataset.  We compare our results with KLT and CoMaL re-detect-and-match, and show superior performance on the boundaries of objects when compared to KLT and an overall improvement when compared with CoMaL re-detect-and-match in almost all cases.   This can potentially improve the performance of existing vehicle tracking systems which use KLT~\cite{song2014vehicle, tanathong2014translation, Sinha2011}. Finally, we evaluate our algorithm on the domain of Human Tracking. Using point trajectories has been a common theme in several Action Recognition algorithms.  We show that the overall performance of our algorithm is better than KLT on the human tracking dataset.  Thus, our algorithm can potentially improve the performance of several Action Recognition algorithms that rely on the  KLT~\cite{wang2011action,messing2009activity,matikainen2009trajectons, uemura2008feature, sun2010activity}.

\subsubsection{Object Tracking on the CoMaL Dataset}

\begin{figure}[t]
\centering
\includegraphics[width=1\linewidth, trim=0 185 150 0, clip]{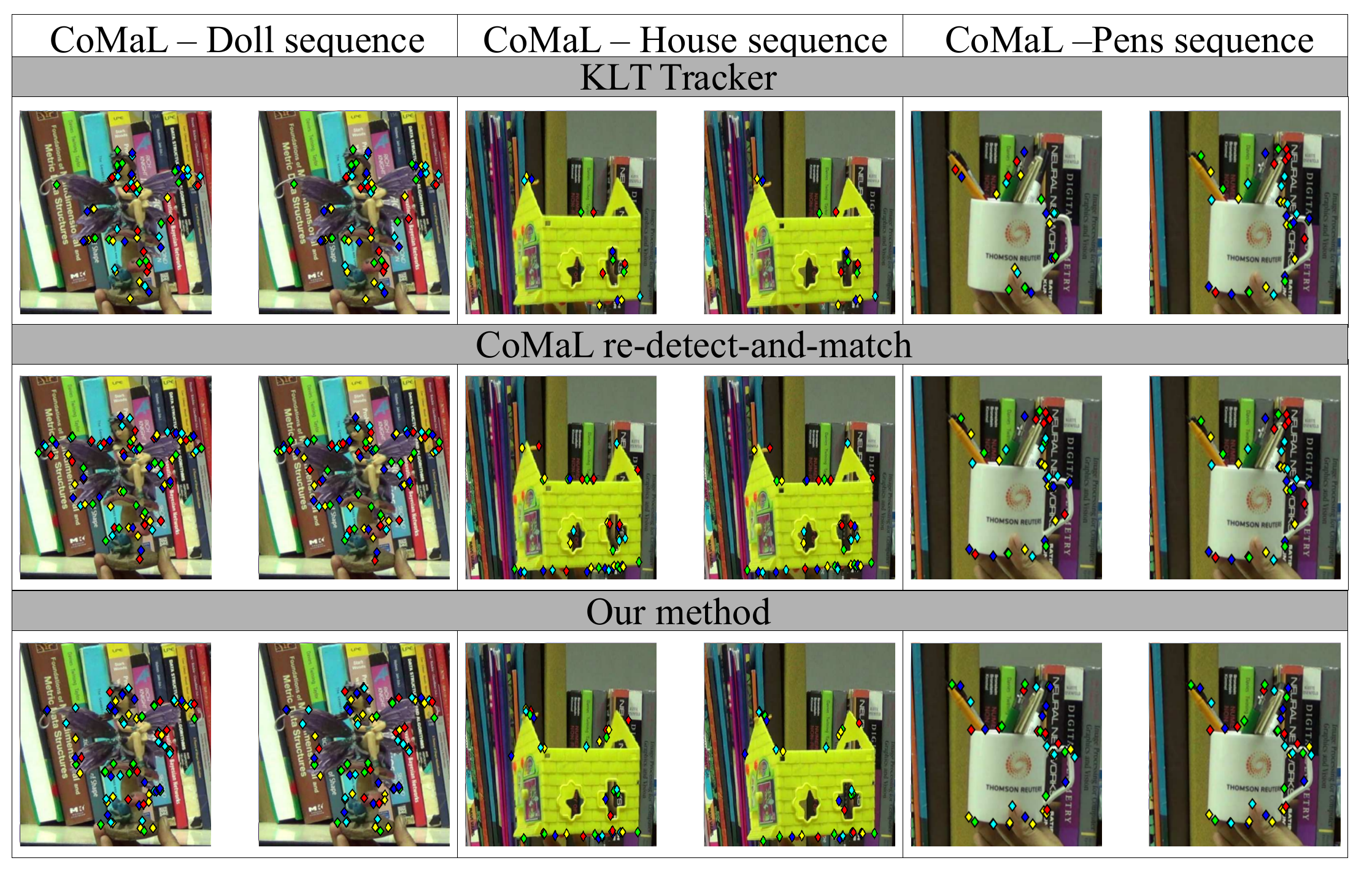}
\caption{Qualitative results on the boundaries for 3 CoMaL sequences.}
\label{fig:qualitative_comal}
\end{figure}

This is a controlled setting where we evaluate CoMaL redetect-and-match, KLT and our tracking algorithms on the dataset provided by the authors of the CoMaL Detector~\cite{Ravindran_2016_CVPR}.  Tracking of feature points at the boundaries is difficult in this dataset due to a large texture in the background of the objects, leading to a large variation in the support region of the boundary points. The dataset provides images for the background in order to perform background subtraction and obtain the foreground pixels.  Thus we can compute the boundary regions which enables the evaluation the different methods on the boundary and non-boundary regions separately.  Some qualitative results are shown in Figure~\ref{fig:qualitative_comal} while 
Table~\ref{table:comal_results} shows some quantitative results.  Our tracking algorithm clearly outperforms KLT on the boundary regions as expected. The background portion of the support region changes very frequently in the CoMaL dataset, due to which KLT cannot track effectively. However, the tracker slightly underperforms compared to CoMaL re-detect-and-match on the boundary regions. We improve over the re-detect-and-match in the interior regions as expected. We also observe a slight improvement over KLT in the interior regions possibly because of the more stable nature of the level line approach compared to a patch matching approach. 

\begin{table}[t]
    \centering
    \scalebox{0.6}{
    \begin{tabular}{|c|c|c|c|c|c|c|}
        \hline
		Sequence & Doll & Hero & House & Toy & Pens & Race-car\\
		\hline
        \hline
		Method & \multicolumn{6}{|c|}{Boundary Region} \\ \hline
		CoMaL TD & \textbf{80.6/1.0} & \textbf{76.8/1.0} & \textbf{53.0/1.0} & \textbf{55.5/1.0} & \textbf{54.1/1.0} & \textbf{85.7/1.0}\\
		KLT & 43.2/1.0 & 40.5/1.0 & 32.6/1.0 & 28.5/1.0 & 22.4/1.0 & 40.1/1.0 \\
		Ours & 68.5/1.0 & 72.2/1.0 & 40.0/1.0 & 34.1/1.0 & 48.7/1.0 & 73.1/1.0 \\ \hline\hline
		Method & \multicolumn{6}{|c|}{Non-Boundary Region} \\ \hline
		CoMaL TD & 143.0/1.0 & 201.8/1.0 & 91.7/0.9 & 108.8/1.0 & 70.3/1.0 & 138.3/1.0 \\
		KLT & 138.6/1.0 & 156.2/1.0 & \textbf{178.5/0.9} & 95.3/1.0 & 71.9/1.0 & 136.0/1.0 \\
		Ours & \textbf{203.6/1.0} & \textbf{254.3/1.0} & 138.2/0.9 & \textbf{142.5/1.0} & \textbf{104.9/1.0} & \textbf{205.2/1.0} \\
		\hline
    \end{tabular}
    }   
    \vspace{0.1cm}
	\caption{Results on CoMaL dataset on the boundary and non-boundary regions.}
    \label{table:comal_results}
\end{table}

\subsubsection{Vehicle Tracking}

\begin{figure}[t]
\centering
\includegraphics[width=1.0\linewidth, trim=0 230 60 5, clip]{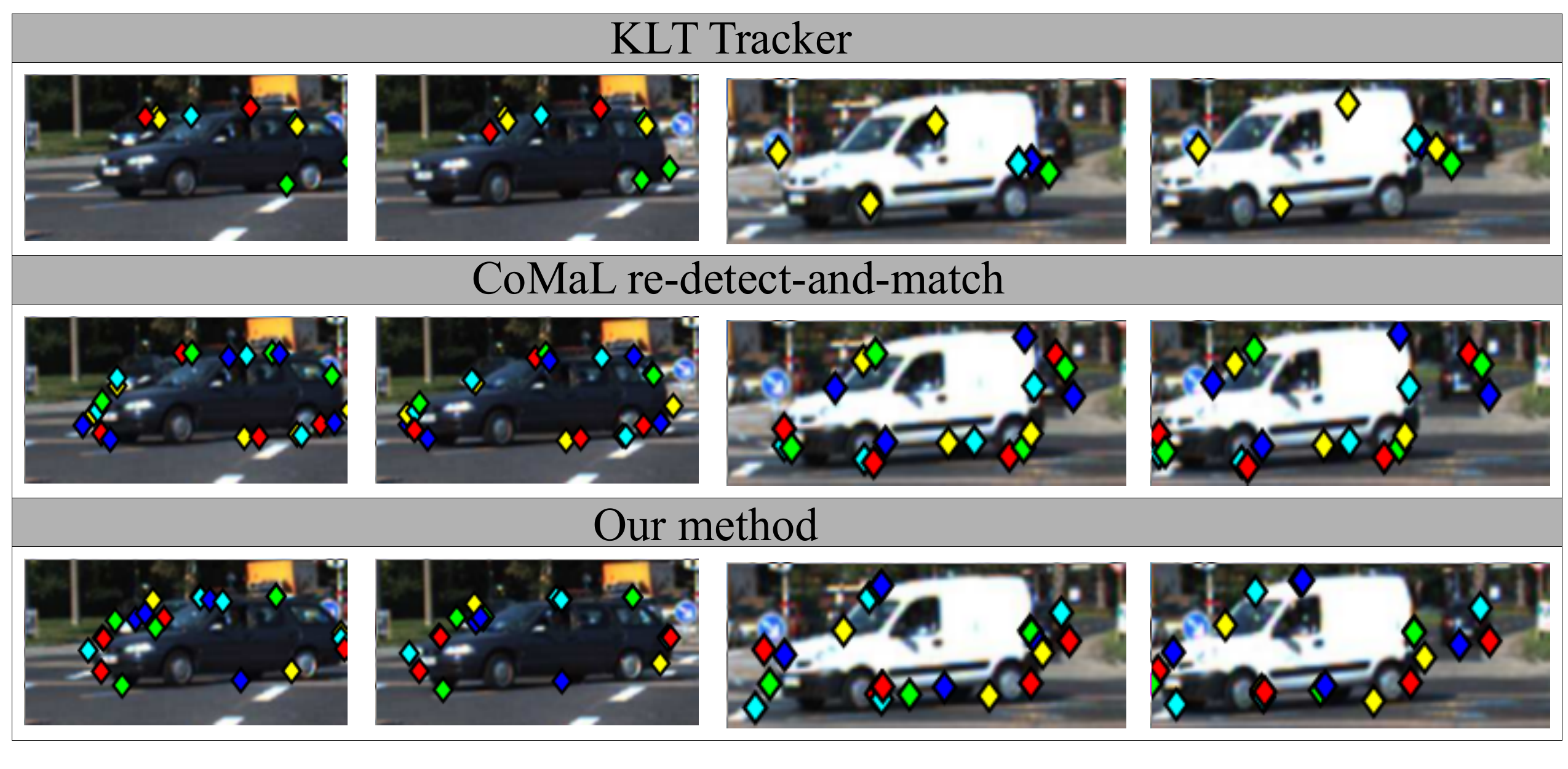}
\caption{Qualitative results on the boundaries for CarC sequence of KITTI dataset.}
\label{fig:qualitative_kitti}
\end{figure}

\begin{table}[t]
    \centering
    \scalebox{0.7}{
    \begin{tabular}{|c|c|c|c|c|}
        \hline
		Sequence & CarA & CarC & CarF& CarG \\
		\hline
		\hline
		Method & \multicolumn{4}{c|}{Boundary Region} \\ \hline
		KLT & 48.4/0.7 & 52.9/0.8 & 26.2/0.9 & 51.7/0.7 \\
		CoMaL TBD & \textbf{76.0/0.8} & 86.2/0.8  & 33.9/0.9 & 64.2/0.7 \\
        Ours & 75.9/0.7 & \textbf{91.0/0.8} & \textbf{51.8/0.9} & \textbf{71.7/0.7}\\
		\hline
		Method & \multicolumn{4}{c|}{Non-Boundary Region} \\ \hline
		KLT & 174.0/0.7 & 263.8/0.8 & 97.9/0.9 & 168.0/0.7\\
        CoMaL TBD & 127.2/0.7 & 204.0/0.8 & 33.9/0.9 & 148.3/0.7 \\
		Ours & \textbf{193.7/0.7} & \textbf{326.7/0.8} & \textbf{141.2/0.9} & \textbf{225.9/0.7}\\
		\hline
		Method & \multicolumn{4}{c|}{Overall} \\ \hline
		KLT & 222.4/0.7 & 316.8/0.8 & 124.2/0.9 & 219.7/0.7\\
        CoMaL TBD & 203.2/0.7 & 290.1/0.8 & 64.9/0.9 & 212.6/0.7\\
		Ours & \textbf{270.4/0.7} & \textbf{417.6/0.8} & \textbf{193.0/0.9} & \textbf{297.6/0.7}\\
		\hline
    \end{tabular}
    }   
    \vspace{0.1cm}
	\caption{Boundary, Non-Boundary regions and Overall results on four sequences of the KITTI dataset.}
    \label{table:kitti_results}
\end{table}

Next, we evaluate our algorithm on the vehicle tracking problem.  Point tracking has been applied extensively for this application~\cite{jodoin2014urban, nister2004visual, pollefeys2008detailed, sakurada2013detecting, song2014vehicle, tanathong2014translation}, where KLT~\cite{lucas1981iterative} is the most common choice~\cite{song2014vehicle, tanathong2014translation, Sinha2011, nister2004visual, pollefeys2008detailed}. Vehicle tracking has become increasingly important with the impending advent of autonomous vehicles, traffic surveillance systems, \etc.  Since it is possible for vehicles to have uniform surfaces hindering the use of corners, edges, \etc in the interior of the vehicle, it is important to fully utilize the boundary information of the vehicle for optimum tracking and hence, the performance of the point trackers at the boundaries is important (A recent crash of a Tesla car due to a homogeneous tractor trailer is a relevant case). 
Also, while learning-based vehicle tracking algorithms have been fairly successful at the task of object tracking recently, these algorithms might fail when the object is not fully visible in the image or if an object or variations in pose of objects which were unseen in the training data are observed in the scene. This necessitates augmentation of such learning-based approaches with conventional feature point-based approaches in order to make the systems more robust.    To test the efficacy of our tracker, we evaluate the performance on 4 sequences of the KITTI dataset~\cite{Geiger2012CVPR}. The remaining sequences had relatively low frame-rates which hinder the performance of any point based tracking algorithm, so we do not report results on them.  In the KITTI dataset, video sequences are taken from moving vehicles and present realistic scenarios for autonomous driving.  The results obtained are shown in Table~\ref{table:kitti_results}.  In order to provide a comparison on the boundary and interior regions, we segmented the vehicles in the 4 sequences by manually providing interactive inputs to the GrabCut~\cite{rother2004grabcut} algorithm.  We obtained a segmentation for all the frames of the CarA sequence, and only 100 consecutive frames in CarC, CarF and CarG sequences, as this was a manual effort and hence time-consuming.  Our tracking method outperforms KLT by a significant margin on the boundaries in all the four sequences. It also slightly improves over CoMaL re-detect-and-match  on the boundaries as well as the interior.  Thus, our algorithm works as expected in the realistic scenario of autonomous driving as presented by the KITTI dataset. 
Some qualitative results are provided in Figure~\ref{fig:qualitative_kitti}.

\begin{table*}[t]
    \centering
    \scalebox{0.8}{
    \begin{tabular}{|c|c|c|c|c|c|c|c|c|}
        \hline
		Sequence & Frame rate & MOT-02 & MOT-04 & MOT-05 & MOT-09 & MOT-10 & MOT-11 & MOT-13\\
        \hline\hline
		\multirow{3}{*}{KLT} & $Original$ & 91.6/0.6 & 102.0/0.8 & \textbf{715.9/0.7} & 64.5/0.8 & 48.8/0.7 & 531.0/0.8 & 226.7/0.6 \\
		& $\frac{Original}{2}$ & 97.1/0.6 & 101.8/0.8 & 311.0/0.7 & 45.9/0.8 & 22.4/0.7 & 433.5/0.8 & 184.3/0.6 \\
		
		& $\frac{Original}{4}$ & 81.9/0.6 & 92.1/0.8 & 363.4/0.6 & 15.8/0.8 & 7.1/0.7 & 185.8/0.8 & 132.4/0.6 \\
		\hline\hline
		\multirow{3}{*}{Ours} & $Original$ & \textbf{94.8/0.6} & \textbf{191.0/0.8} & 656.6/0.7 & \textbf{136.0/0.8} & \textbf{83.5/0.8} & \textbf{621.5/0.8} & \textbf{254.2/0.6} \\
    	& $\frac{Original}{2}$ & 83.2/0.6 & 188.7/0.8 & 443.0/0.7 & 97.4/0.8 & 65.7/0.8 & 495.7/0.8 & 207.8/0.6 \\
		   & $\frac{Original}{4}$ & 67.0/0.6 & 175.6/0.8 & 396.3/0.6 & 55.3/0.8 & 44.7/0.8 & 361.5/0.8 & 153.7/0.6 \\ 
		\hline
	\end{tabular}
    }   
    \vspace{0.1cm}
	\caption{Results on the different sequences of the MOT 2016 training dataset at different frame-rates. $Original$ refers to the original frame-rate. $\frac{Original}{n}$ refers to the video sequence sampled at every $n^{th}$ frame. }
    \label{table:mot_results}
\end{table*}

\subsubsection{Human Tracking}

\begin{figure}[h]
\centering
\includegraphics[width=1.0\linewidth, trim=70 150 300 30, clip]{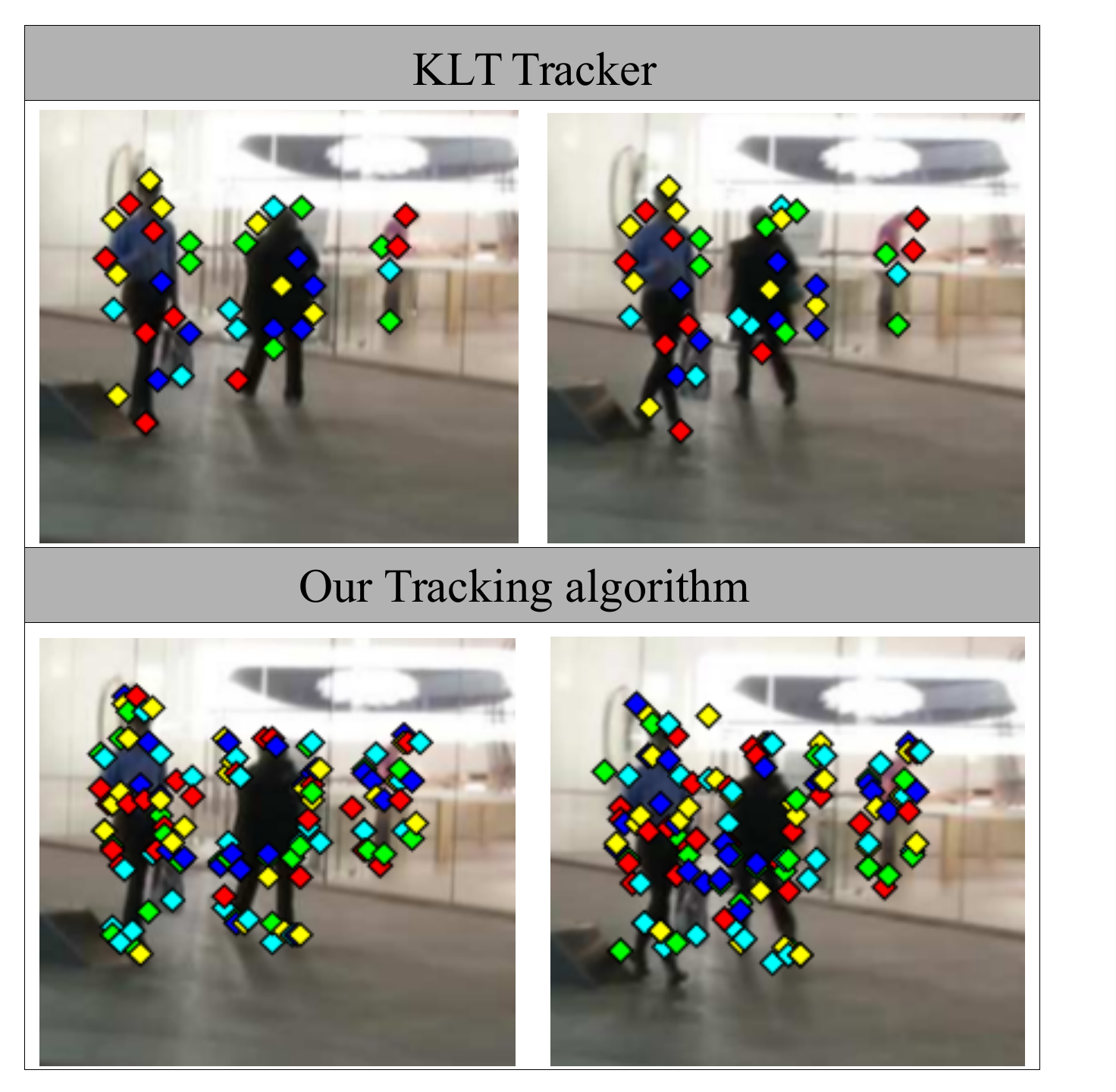}
\caption{Overall tracking results for image patches from two consecutive frames in the MOT-10 sequence.}
\label{fig:qualitative_mot}
\end{figure}

Point Tracking has been used extensively in the Action Recognition community where KLT is again the most popular choice for obtaining point trajectory-based features~\cite{wang2011action,messing2009activity,matikainen2009trajectons, uemura2008feature, sun2010activity}. We show the efficacy of our tracker for the domain of Human Tracking. Results are shown on the MOT 2016 challenge training video sequences.  The dataset provides ground truth trajectories for the bounding boxes of humans in the scene in the training set, which we use to generate the ground truth for point tracking.  The sequences are challenging because the cameras are moving and include crowded scenes such as shopping malls, busy streets, \etc. Since the frame-rate of the provided sequences was high, we show variations in the performance of our tracking algorithm as the frame rate reduces.  For KLT, this can be more of a challenge as it uses a gradient-descent approach.  For us, this can also reduce the performance since one will have to search in a larger window, which can increase the running time, apart from increasing the chances of a wrong match.  As outlined in Section~\ref{sec:evaluation}, we generate the ground truth by assuming that the relative location of the points \wrt the annotated bounding box remains the same.  Some qualitative results are shown in Figure~\ref{fig:qualitative_mot} while the quantitative results are shown in Table~\ref{table:mot_results}.  While we show KLT's variation in performance with the frame rates, it may not be a fair comparison as the KLT tracker was not designed to work for low frame rates. Our algorithm outperforms KLT at the original frame rate on six out of the seven sequences.  We can also observe that our performance does not drop significantly as the frame rate decreases, which is expected because of our robust two-stage tracking, although we had to increase the search window in the case of lower frame rates due to a higher object motion.  Due to unavailability of segmentation information and infeasibility of annotating the dataset, we report only the overall results and do not have the boundary and non-boundary classification of the points.  

\section{Conclusions and Future Work}

We have proposed an accurate tracking algorithm for tracking Feature points on the Object Boundaries. We track the CoMaL Feature points which are shown to be superior for detection and matching on object boundaries. This is achieved by first tracking the level line segment associated with the corner by matching it with level lines obtained in the next frame using MSER detection. The level lines are initially matched using the Hierarchical Chamfer Matching Algorithm to filter out poor matches, and the shortlisted matches are then verified using Part SSD matching to obtain the best match.  Tracking results on three different scenarios of Object Tracking, Vehicle Tracking and Human Tracking show significant improvement in performance at the object boundaries when compared to the current state-of-the-art in tracking, \ie KLT, and also an overall improvement in performance compared to the CoMaL re-detect-and-match framework proposed earlier. It is also more efficient than the CoMaL re-detect-and-match framework. 

Future work includes development of a real-time implementation of the tracker, possibly by utilization of GPUs.

{\small
\bibliographystyle{ieee}
\bibliography{comal_tracking_cvpr2017.bib}
}

\end{document}